\documentclass[letterpaper, 10 pt, conference]{ieeeconf}  %

\IEEEoverridecommandlockouts                              %

\usepackage{graphics} %
\usepackage{epsfig} %
\usepackage{mathptmx} %
\usepackage{times} %
\usepackage{amsmath} %
\usepackage{amssymb}    %
\usepackage{bm}
\usepackage{algorithm}
\usepackage{multirow}
\usepackage{verbatim}
\usepackage{graphicx}
\usepackage{subfigure}
\usepackage{color}
\usepackage{url}
\usepackage[font=small,labelfont=bf]{caption}
\usepackage{arydshln}
\usepackage{hyperref}

\def \kkk{\color{black}}

\title{\LARGE \bf
Online Object and Task Learning via Human Robot Interaction 
}

\author{Masood Dehghan$^{*}$, Zichen Zhang$^{*}$, Mennatullah Siam$^{*}$, Jun Jin, Laura Petrich and Martin Jagersand %
\thanks{All authors are with the Department of Computing Science, University of Alberta, Canada.
\{masood1, \,zichen2, \,mennatul, \,jjin5, \,llorrain, \,mj7\}@ualberta.ca}
\thanks{* Equal contributions}%
}

\begin{document}

\maketitle
\thispagestyle{empty}
\pagestyle{empty}

\begin{abstract}
This work describes the development of a robotic system that acquires knowledge incrementally through human interaction where new objects and motions are taught on the fly.
The robotic system developed was one of the five finalists in the KUKA Innovation Award competition and demonstrated during the Hanover Messe 2018 in Germany.  
The main contributions of the system are 
i) a novel incremental object learning module - a deep learning based localization  and recognition system - that allows a human to teach new objects to the robot,
ii) an intuitive user interface for specifying 3D motion task associated with the new object, and
iii) a hybrid force-vision control module for  performing compliant motion on an unstructured surface.
This paper describes the implementation and integration of the main modules of the system and summarizes the lessons learned from the competition.

\end{abstract}

\section{Introduction}
        \label{sec:introduction}
        A key challenge in deploying robots in human environments is the uncertainty and ever-changing nature of  the  human environment.  To adapt to the variability, the robot needs to constantly update its model of the world. In other words, the robot needs to be capable of learning incrementally. One of the first steps of interacting with the world is to recognize new objects and know how to utilize them. This is a difficult problem to solve. One approach to tackle this problem is to leverage human's knowledge and guidance whenever the robot is not able to  make a decision on its own \cite{leeper2012strategies}.

One of the earliest attempts toward incrementally learning novel objects for robot manipulation was demonstrated in \cite{ude2008making}. An interactive system that places the objects in the robot's hand to be learned, in order for the robot to perform acquisition of the different object views. The method utilized hand crafted features with Gabor filters for constructing the objects representation. In \cite{krainin2011manipulator} a method was proposed that tracks the object and a robotic manipulator, while constructing a 3D model for the object based on surfels. In \cite{bevec2013object} the robot controls its gaze based on the detected objects locations in order to collect further poses of each object. In \cite{pasquale2015real} the ICUB World dataset was proposed that focuses on incrementally learning object recognition and detection of novel objects using a human robot interaction approach. Different deep learning methods were proposed \cite{camoriano2017incremental} \cite{pasquale2016object} and evaluated on the ICUB World dataset. Along this direction, in our previous work \cite{Valipour17} we  proposed a method to improve the robot's visual perception incrementally and used human robot interaction (HRI) to learn new objects and correct false interpretations.

\begin{figure}[t]
\centering
\includegraphics[width=.48\textwidth] {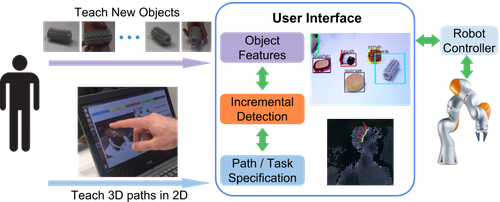}
\caption{System overview: (Top left) the image sequence of comb automatically cropped by the system to extract features of the new object.
(bottom left) the user draws desired combing strokes on a touch-screen. (Right) User interface has three modules: \textit{Path Specification} receives 2D combing strokes, projects them on the 3D head surface and construct the 3D path points; \textit{Incremental Detection} performs object detection on incrementally added new object classes; \textit{Control Module} enables the robot to perform compliant 3D combing motion while maintaining the contact with the hair.} 
\label{fig:system_overview}
\end{figure}

In this paper, we build on the same idea and develop a new system with a learning module that is more natural and efficient, and allows teaching tasks that are associated with the new objects. 
In particular, the new system improves on the following aspects: 1) when teaching a new object to the robot, it does not require that the object remains static anymore. It automatically tracks the object such that a human can teach a new object just by holding it in hand and showing it to the camera,
2) a novel incremental object detection system that has robust performance on newly learned objects, existing objects, and objects from unknown classes, and supports open-set recognition. 
3) an intuitive user interface for specifying 3D motion task associated with the new object, and
4) a hybrid vision-force control for performing compliant motions that require  contact with unstructured surfaces. 

Our developed robotic system was one of the top five finalists in the  KUKA Innovation Award competition 2018 \cite{KUKA_IA2018}.
This year's theme of the competition was ``Real-World Interaction Challenge''.  
The competition aimed to seek robotics solutions that adapted to the changing environment in the real world. We performed live demos at the Hannover Messe 2018. In these demos the audience brought new objects. Our robot system learned both the visual appearance of the new objects and how to use them in contact motions w.r.t. a sensed surface.

\kkk

The rest of the paper is organized as follows. 
Section \ref{sec:system_description} outlines the overall system modules and their interconnections.
The object localization module is described in section \ref{sec:localization}. The details of the proposed incremental classification module is presented in Sec \ref{sec:incremental} followed by the user interface design in Sec \ref{sec:user_interface}. Details of the hybrid force-vision control module is explained in Sec \ref{sec:control}. Experimental results and the demonstration at the Hannover Messe are presented in Sec \ref{sec:experiments}.
Section \ref{sec:conclusions} concludes the paper and summarizes the lessons learned during the competition.

\section{System Overview}
        \label{sec:system_description}

The overall architecture of our system is illustrated in Fig.~\ref{fig:system_overview}.
It learns new objects and tasks for contact motions on unstructured surfaces, in an interactive way.
An example of a use case is to teach the robot how to comb hair, which requires teaching what is a comb, where to move the comb and how to properly align with the head during the movement.
With our system, a user can teach the robot what a comb is just like teaching another human. The user only needs to hold a comb in his/her hand and rotate it to show different object poses to the camera. The learning module automatically tracks the object and stores the features. From there, the incremental detection module will be able to detect the comb and picks it up from the table. The user can then teach the movement trajectory by drawing paths on a touch-screen.

Our hardware consists of a KUKA LBR iiwa arm instrumented with a flexible Festo gripper \cite{LBRgrippers}, a  pointgrey camera (for teaching objects), two RGB-D sensors and a touch-screen for user interaction. 
Our system  is composed of three main modules, all of which are fully integrated with ROS~\cite{quigley2009ros}:
\begin{itemize}
    \item \textbf{Incremental Object Detection}: an object detection system that allows incrementally adding new classes. We dubbed it as ``Incremental'' in contrast with the traditional method. We adopt a two-stage approach, object localization followed by incremental classification; 
\item \textbf{Visual User Interface} for user interaction: enables the operator to seamlessly add/remove object classes, and the associated paths. The user defines the paths by drawing 2D trajectories on the touch-screen. 
\item \textbf{Hybrid force-vision control}: this module enables the robot to perform 3D motion tasks that requires contact with unstructured surfaces. It receives the RGBD sensor information and constructs the 3D motion trajectories. 
\end{itemize}

The details of the modules are described next. 

\section{Object Localization}
\label{sec:localization}
The objects are placed on a table in front of the robot manipulator. 
Before the objects can be classified into different categories, the robot needs to localize them. We adopt the Region Proposal Network from Faster RCNN \cite{ren2015faster} for the object localization. 
This network serves as a generic object detector, that predicts bounding boxes of the objects in an image and the associated objectness.
Instead of using the downstream classification network layers in Faster RCNN, we use the incremental classification method described in Section~\ref{sec:incremental} such that we can incrementally add new classes.
We use VGG16 \cite{vgg16} as the backbone network, trained on MS-COCO \cite{lin2014microsoft} dataset.  The image patches inside each predicted bounding box are passed as the input to the incremental classification stage. The location of each bounding box is sent to the path specification module.
We optimize the speed of the localization by only processing the latest image while skipping all the images that are observed during the last network inference.

We have tried other alternatives, like using a one-stage object detector YOLOv2~\cite{yolo9000}.
In contrast to the two-stage approach in Faster RCNN, YOLOv2 predicts the objectness and class simultaneously for every anchor box, which comes from priors learned from the training dataset by clustering.
In our setting, we keep the anchor boxes that had high objectnesses as the set of potential objects. For each of these objects, we pass them as input to the incremental classification stage and update its class label according to the classification result.
YOLOv2 achieved great balance of speed and performance on PASCAL \cite{pascal-voc-2012} and COCO datasets. 
However, it did not perform as well as the Faster RCNN based approach in our task. The main reason is that in YOLOv2, while the priors of the anchors learned from these datasets may be a good representation of object locations in the test set, they do not necessarily fit the real-world scenarios.
The user may place an object anywhere on the table. If it happens to be in a location less represented in the dataset, it will get a lower objectness and may be considered a non-object.
On the contrary, in Fastetr RCNN, the anchors are uniformly distributed across the image so it has a better chance of localizing objects in rare locations.

\section{Incremental classification}
\label{sec:incremental}

One approach for object classification is to depend on large-scale a-priori training data. However, large-scale training datasets such as ImageNet \cite{deng2009imagenet} or MSCOCO do not contain all the objects and tools used in a variety of manipulation tasks. 
Nonetheless, the learned convolutional features from pre-training on large-scale training data can benefit from the incremental learning of new objects. 
Unlike the general classification problems where one image is required to be classified based on previous training data that do not capture the object poses, the classification in a human robot interaction setting has additional temporal information of the object undergoing different transformations. 
The different object poses can aid in building better classification modules and greatly benefit the low shot recognition problem. 
Finally, to enable the robot to operate in different environments and incrementally learn objects, it needs to acknowledge when an object presented is from an unknown class. The ability to recognize objects outside of the closed set in its own data is termed as the open-set recognition problem \cite{bendale2016towards}, and allows the robot to request the human help.

Our classification module is comprised of two stages: (1) Teaching Phase, (2) Inference Phase. 
During the teaching phase a human demonstrates different object poses, while during the inference phase the robot is required to detect the novel objects that it has learned. 
This mimics children being taught about novel objects by a teacher or parent  \cite{markman1989categorization}. 

\begin{figure}[tb]
\centering
    \includegraphics[trim=0mm 20mm 0mm 0mm, clip=true, width=0.49\textwidth]{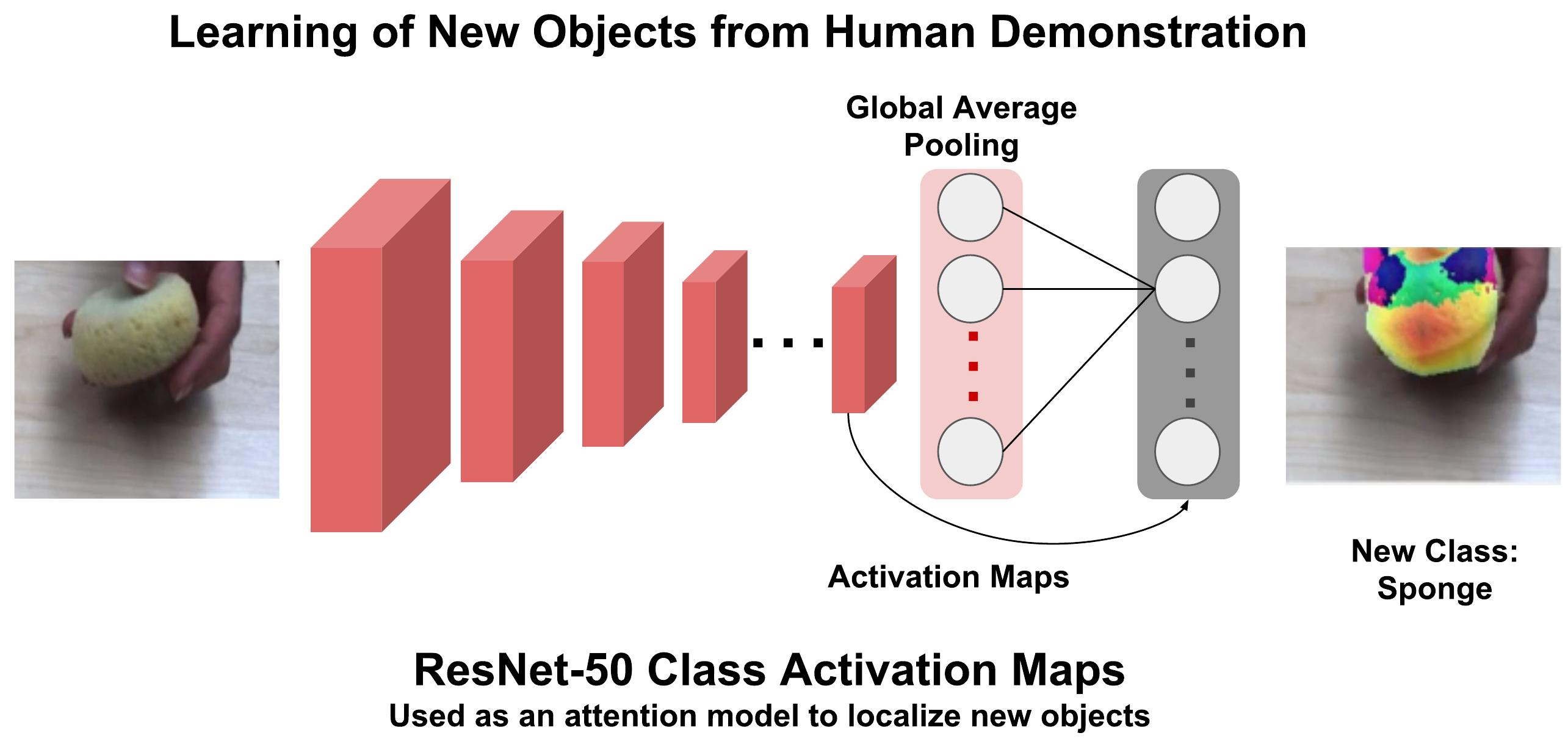}
    \caption{Teaching a new object}
     \label{fig:object_teaching}
\end{figure}

\begin{figure}[tb]
\centering
    \includegraphics[width=0.5\textwidth]{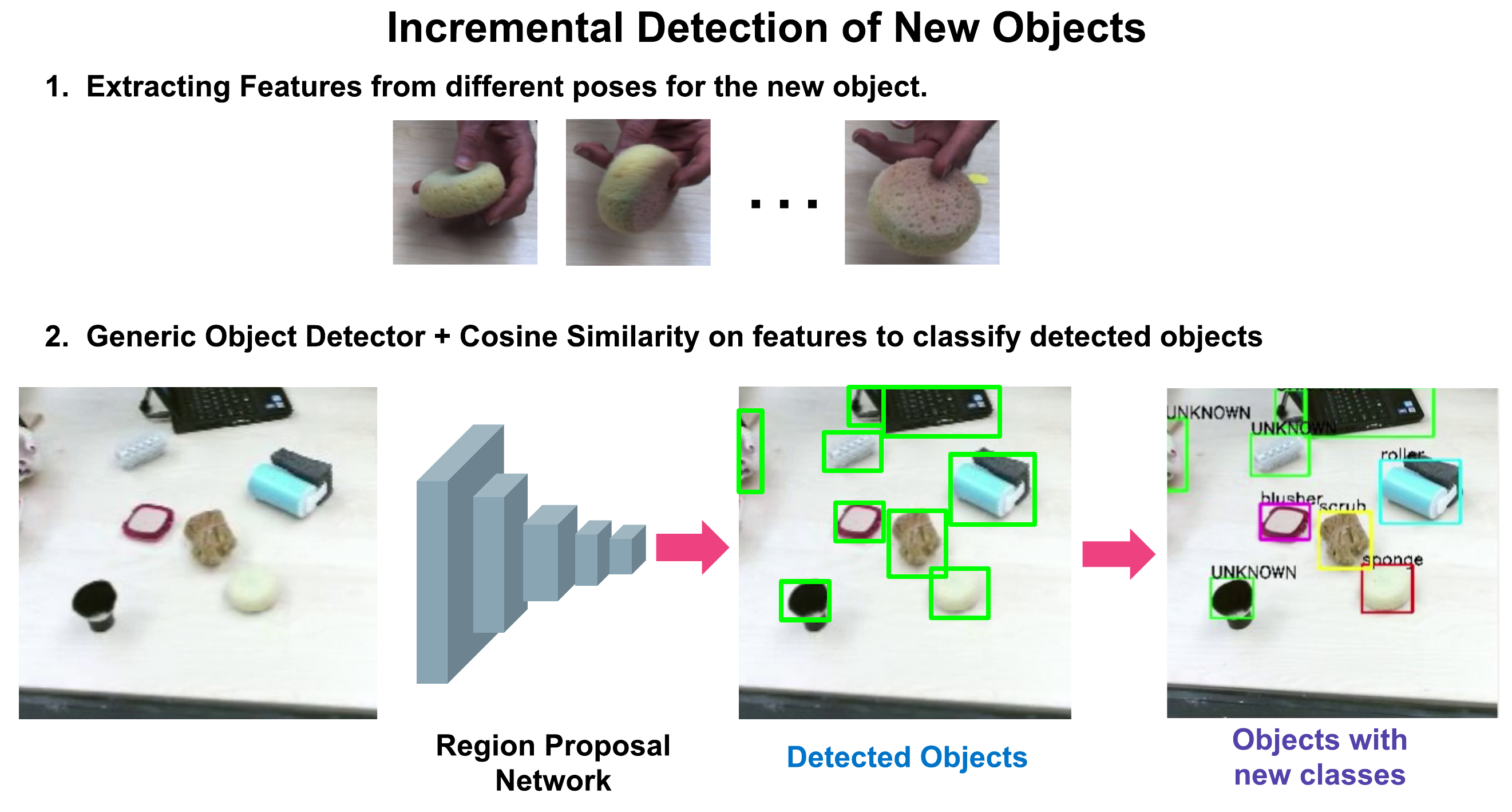}
    \caption{Overview of the incremental object classification.}
     \label{fig:detection-overview}
\end{figure}

In the teaching phase a saliency method based on class activation maps (CAM) \cite{zhou2016learning} from ResNet-50 \cite{he2016deep} is used to automatically detect the object being demonstrated. Although the learned object might be a novel one that does not belong to ImageNet set of classes, CAM will still be able to correctly localize the salient object as shown in Figure \ref{fig:object_teaching}. 
The features from a ResNet-50 network pretrained on ImageNet are extracted for the image patches extracted based on CAM localization. The mean of activations as shown in a previous few-shot learning study \cite{qiao2017few} acts as a strong indicator for the object class. Thus the mean of activations from ResNet-50 for all object poses are used to represent each object. Each novel object being taught to the robot creates a new set of features and its corresponding mean. 

During the inference phase, image patches based on the computed bounding boxes from the object localization module are extracted and their corresponding ResNet-50 features are computed. These act as the query features corresponding to the query objects, and the classification problem is dealt with as a retrieval problem. The query object is classified based on the nearest neighbour of the query feature.
Nearest neighbour algorithm is used in our application since it requires open-set recognition: not only that we need to classify the objects, we also need to know when an object is from an unknown class.
To do that, a distance ratio is computed between the first and second nearest neighbour distances. 
A higher ratio near 1 indicates an ambiguous classification and is rather classified as an \textit{Unknown} object. While a lower ratio indicates higher discrimination between the first and second nearest classes, and the nearest neighbour class is used. 
An overview of the two phases are shown in Figure \ref{fig:detection-overview}.

\section{User interface}
\label{sec:user_interface}

\begin{figure}[ht!]
\centering
\includegraphics[width=.48\textwidth] {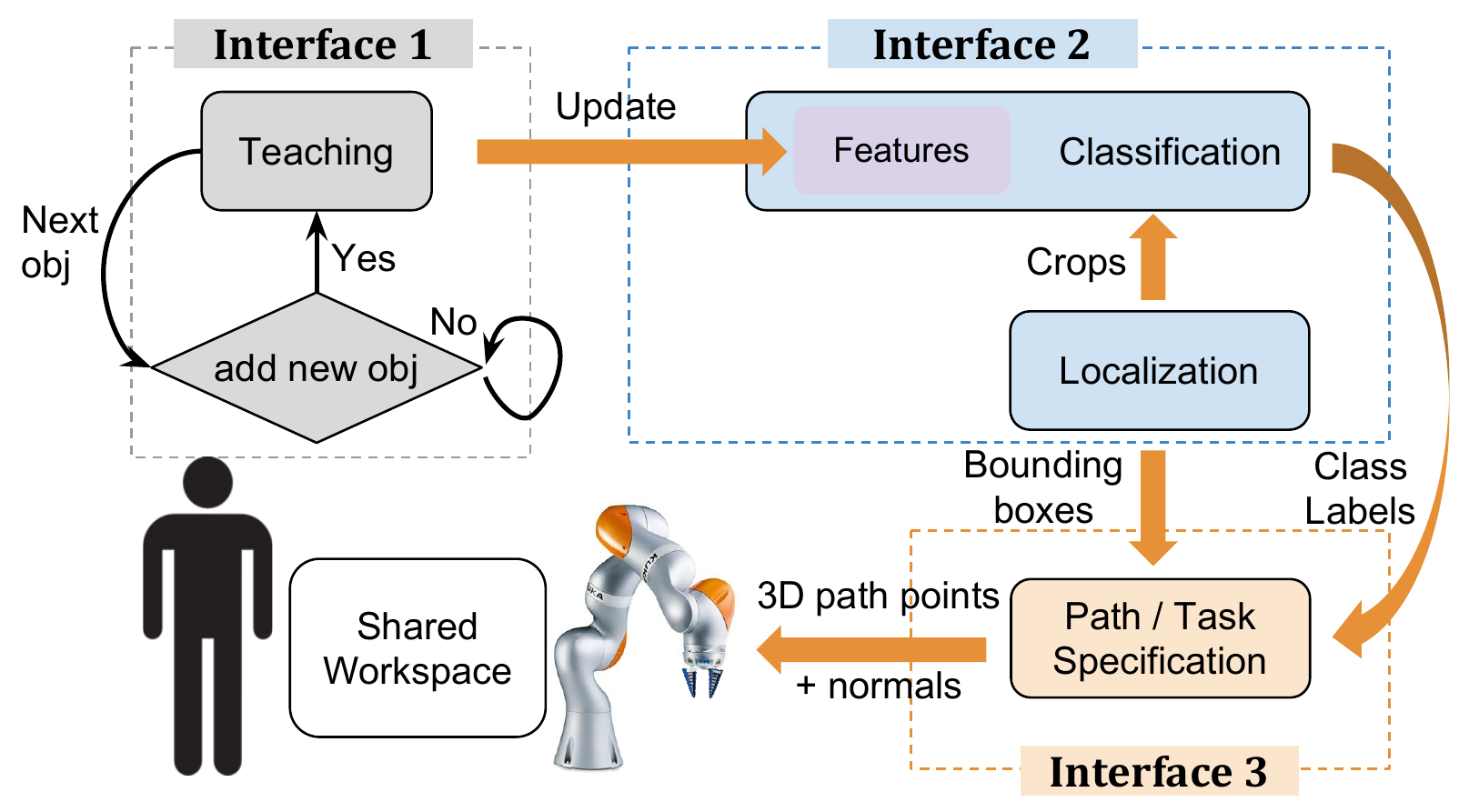}
\caption{User interface modules and the workflow}
\label{fig:workflow}
\end{figure}

The user interface and the workflow is depicted in Fig. \ref{fig:workflow}.
At the start of the system, the user decides whether to add a new class or not. If yes, the user needs to show the new object to the camera along with entering the name of the class. The system will collect samples of this new object and move on to the incremental object detection module. All the objects placed on the table in front of the robot will be detected and assigned as either one of the old classes, or the new class, or an unknown class.
If the user choose not to add a new object, the system can either wait, or bypass the learning module and detect the object as one of the old classes or unknown.
The system allows the user to specify a 3D motion by drawing on the touch screen, and pair this motion with one of the objects.
Jin et al.~\cite{Jin2018RobotEC} attempted visual path specification by watching human demonstrations. In our approach, we allow specifying a more accurate path while still making it feels natural to a human operator.
The robotic manipulator will automatically pick up the object and apply the user-defined motion. The interface is implemented with the qt-ROS package.

\begin{figure}[t]
\centering
\subfigure[Teaching Pane]
{\includegraphics[trim=0mm 0mm 0mm 5mm, clip=true, width=.22\textwidth] {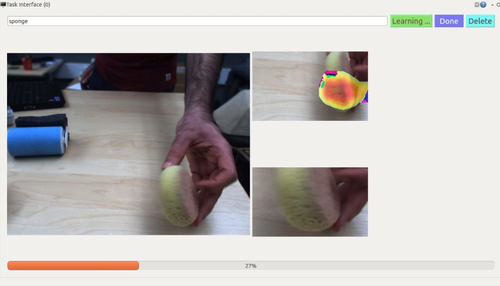}} \hfill
\subfigure[Detection Pane]
{\includegraphics[trim=0mm 0mm 0mm 5mm, clip=true, width=.22\textwidth] {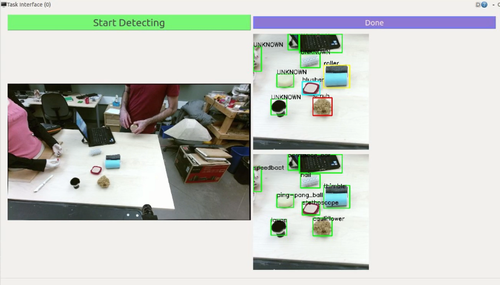}}
\subfigure[Task Pane]
{\includegraphics[trim=0mm 0mm 0mm 5mm, clip=true, width=.22\textwidth] {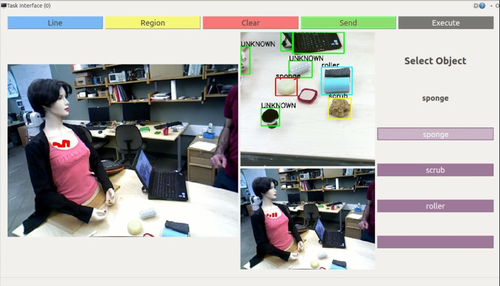}}
\caption{The three panes of the User Interface  a) Teaching pane, where the user teaches the robot new classes of objects; b) Detection pane, where different classes of objects are localized, classified and highlighted to the user;
c) Path specification pane, where the user is able to specify the path by drawing on the touch-screen.}
\label{fig:visual_interface}
\end{figure}

The \textit{User Interface} has three view panes as shown in Fig.~\ref{fig:visual_interface}:
\begin{itemize}
\item Teaching Pane: which allows the user to introduce new objects and define the class by entering its class name. 
It subscribes to the cropped images of new object and shows it to the user. The system collects 200 sample images of the new object, stores them and passes them to the classification module.  
Once the image capturing is done, the deep feature extraction process is activated and runs until completion.
\item Detection Pane: This pane subscribes to the scene video stream and detects all the objects in the scene. Detected objects are displayed to the user and the corresponding names and locations of the objects are stored for the next phase which is motion specification and execution.
\item Task Pane: This pane subscribes to the scene video and enables the user to intuitively specify paths by drawing on the 2D image stream from the scene.
Inspired by \cite{quintero2016vision,DiegoHRI2018}, the user is able to add one or multiple paths or delete the specified path. It is also possible to define area tasks, in which the user selects a region. The interface enclosed the area with a polygon and automatically calculates multiple strokes that covers the selected area. This feature is specifically useful for cleaning or polishing tasks of unstructured surfaces. 
We have not implemented the feature of partially modifying a defined trajectory, but it can be easily added later. 
Once satisfied with the path, the user needs to select an object to pair this path with. The robot will attempt to pick up this object and move it along the specified path. The idea is to decouple the object classes from the utility of the object. With this interface, we allow the user to define how to use the object, with the flexibility to pair/unpair the path from the object. 
Note that as the focus of this system is not on grasping, flexible Festo gripper fingers \cite{LBRgrippers}
is used and a simple table-top grasping strategy based on the orientation of the object is used to grasp the objects.  
\end{itemize}

The goal of the controller is to constrain the end-effector motion to the path (on a surface) while maintaining a contact force ($f_n$) on the surface. 
In order to achieve this, we make use of the operational space control. 
Following the work of Khatib \cite{Khatib}, we can decompose the motion task by projecting the motion onto orthogonal directions along the tangential ($ {\bm t}$) and normal ($ {\bm n})$ directions, see Fig. \ref{fig:surface_coordinates}(c). A dedicated controller is then designed for each direction. An additional controller is also required for orienting the robot's tool.  

\section{Hybrid force-vision control Module}
\label{sec:control}

\begin{figure}[t]
\centering
\includegraphics[trim = 0mm 0mm 0mm 0mm, clip=true, width=.45\textwidth]{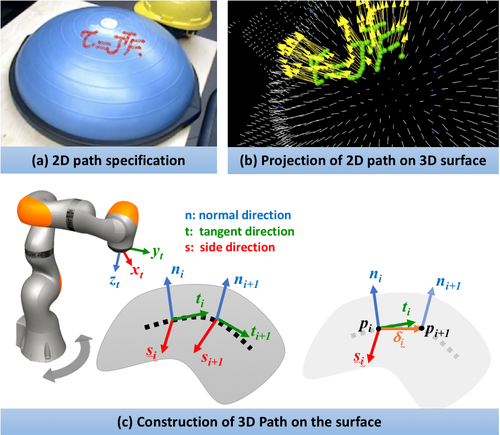}
\caption{Illustration of user-defined 2D path and  generation of the corresponding 3D  path points on the surface.}
\label{fig:surface_coordinates}
\end{figure} 

The overview of the control module is illustrated in Fig. \ref{fig:surface_coordinates}. User-defined 2D paths (drawn  on a touch screen) are first projected on the 3D surface. To do this, we need to compute the  surface normals.

\subsection{Surface Normal Extraction}

This module computes  the 3D path coordinates and the corresponding surface normals. 
The input to the module is a 2D path, which is converted to the 3D path coordinates via the direct correspondence between the RGB sensor and depth sensor. 
The Kinect sensor is calibrated such that these coordinates in the Kinect frame can be transformed into the robot base frame.
To compute the surface normals, we fit a plane to the neighborhood patch $\mathcal P_i$ of the target point $\bm p_i$. 
$\mathcal P_i$ is  obtained from the points within a radius $r$ from $\bm p_i$. 
We can then estimate the target normal ${\bm n}_i$ by computing the smallest eigenvalue $\lambda_{i,0}$ of the covariance matrix ${\bm v}_{i,0}$ of $\mathcal P_i$. The reader is referred to \cite{rusu2013semantic} for the details.

The integral normal estimation  implemented in the PCL library~\cite{rusu20113d} is used. 
It optimizes the computation by taking advantage of the organized structure of the point cloud.
The normal is calculated by taking the  cross product of the two local tangential vectors formed by the right-left pixels and up-down pixels.
Due to the noisy nature of the sensor data, we smooth out the tangential vectors by taking the average, or, an integral image~\cite{holzer2012adaptive}.
In our implementation we use the PCL \textit{Average 3D Gradient} mode which creates 6 integral images to compute smoothed versions of horizontal and vertical 3D gradients~\cite{PerezDehghan2017}.

\subsection{Path Controller} 
It is assumed hereafter that the visual interface provides the desired path, i.e. a down-sampled sequence of 3D points on the surface $\{\bm {p}_0, \cdots,  \bm {p}_i , \cdots  \}$  and the corresponding normal directions at each point $\{\bm {n}_{0}, \cdots,  \bm {n}_{i} , \cdots  \}$.
With the desired path, the tangential direction at each point $\bm t_i$ is computed as follows (see Fig. \ref{fig:surface_coordinates} (c)):    
\begin{align}
    {\bm \delta}_i &= \bm {p_{i+1}} - \bm {p_{i}} \nonumber  \\
    \bm t_i &= \bm \delta_i - ( \bm \delta_i \cdot \bm n_i) \, \bm n_i \nonumber \\
    \bm t_i &= \tfrac{ \bm t_i }{ || \bm t_i ||}  \\
    {\bm s}_i &=  {\bm t}_i \times  {\bm n}_i
\end{align} 
The path reference frame is then fully characterized by 
$ \{ {\bm t} , {\bm n}, {\bm s} \}$. 

The desired orientation of the robot end-effector can also be obtained from the path on the surface.  
In  general, the desired orientation\footnote{We use relative Euler angles with rotation order $\alpha_z$ followed by $\beta_y$ followed by $\gamma_x$.}  $(\alpha_z, \beta_y, \gamma_x)$ could be set path dependant and could change along the the path. For example, suppose that we want to reorient the end-effector (tool) reference frame $\{x_t, y_t, z_t\}$ such that $z_t$ is aligned with $\text{-} {\bm n_i}$ and $y_t$ is aligned with the tangential direction ${\bm t_i}$. This can be achieved by constructing the rotation matrix from tool to base  
$ _t^b{\bm R}_i = \left[
    \begin{array}{c|c|c}
        {\bm s_i} & {\bm t_i} & \text{-} {\bm n_i}
    \end{array}
\right].
$
Following the work of Khatib \cite{Khatib}, the motion task is decomposed onto orthogonal directions along the tangential ($ {\bm t}$) and normal ($ {\bm n})$ directions as shown in Fig. \ref{fig:surface_coordinates}(c). 
Path following in the tangential  
plane (plane constructed by $\{ {\bm s}$,$ {\bm t}\}$) is achieved by designing a compliant controller inspired by  \cite{PerezDehghan2017}.

KUKA LBR iiwa robot is a torque controlled 7-DOF  manipulator with integrated torque sensors at each link.  
Its dynamic equation is of the form 
\begin{align}
\label{eqn:dynamic}
 \bm {M} \ddot {\bm q} + \bm {C} \dot {\bm q} + \bm g  = \bm {\tau} + \bm {\tau}_{ext}
\end{align}
where $\bm q$ is the joint angles, $ \bm {M}$ is the positive-definite inertia matrix, $\bm {C}$ %
is the Coriolis matrix, ${\bm g}$ is the gravitational force, $\bm {\tau}$ the actuators torques, and $\bm {\tau}_{ext}$ is the external generalized force applied to the robot by the environment.\footnote{Note that for simplifying the notations, we drop the dependencies on $\bm q$ unless it is required.}

Assuming that the  end-effector
position and orientation is described by a set
of local coordinates $\bm x \in  \mathbb R^6$ and the forward kinematics map 
$\bm x = f(\bm q$) is known,  the mappings between
joint and Cartesian velocities and accelerations are
    $\dot {\bm x} =       \bm J  \dot {\bm q}$, 
    $\ddot {\bm x} = \dot {\bm J} \dot {\bm q} + \bm J \ddot {\bm q}$, 
where  $\bm J(\bm q) = \frac{\partial f(\bm q)}{\partial (\bm q)}$  is the manipulator Jacobian and has full row rank\footnote{The singular case can be treated using the method described in \cite{Khatib}.}.
Now denote $\bm e_x = {\bm x} - \bm x_d$ as the Cartesian error between actual Cartesian pose ${\bm x}$ and the
desired one $\bm x_d := [\bm p_i^T  , \alpha_z, \beta_y, \gamma_x]^T$.

Our controller sends the torque command: 
  \begin{align}
    \bm \tau &= \bm J^T  \bm F_d + \bm C \dot{\bm q} + \bm g  \label{eqn:tau_imp}\\  
    \bm F_{d} &= \bm M_{\bm x} \ddot{\bm x}_d - \bm D \dot{\bm e}_{\bm x} - \bm K {\bm e}_{\bm x} 
    - {\bm M}_{\bm x} \dot {\bm J} \dot {\bm q} \label{eqn:Fd}
 \end{align}
Using the fact that ${\bm {\tau}}_{ext} = \bm J^T \bm F_{ext} $,  
 $\bm M_{x} = (\bm J {\bm M}^{-1} \bm J^T)^{-1}$ and substituting eqns. (\ref{eqn:tau_imp}) and (\ref{eqn:Fd})  in (\ref{eqn:dynamic}), results in the closed loop system of 
\begin{align}
    \bm M_{x} \, \ddot {\bm e}_{x} + {\bm  D} \, \dot {\bm {e}}_{x}  +  \bm K \, {\bm e}_{x} = \bm F_{ext} 
\end{align}
which has a desired compliance behavior in the presence of external forces and torques at the end-effector $\bm F_{ext} \in \mathbb R^6$. Matrices $\bm  K$ and $\bm  D$ are diagonal and specify the desired stiffness and damping in each direction.  
The stiffness gains in the tangential plane are set high while the stiffness in the normal direction is set to be low.

To maintain the contact with the surface, we need additional  force ($f_n$) in the  normal direction ($-\bm n_i$).
\begin{align}
&\bm F_{n} = - \bm K_{p} \big( (\bm f - \bm f_n) \cdot \bm {n}_i \big) -  \bm K_{d} (\bm {\dot f} \cdot \bm {n}_i) \\
&\bm \tau_n =  \bm J^T  \bm F_n \label{eqn:tau_n}
\end{align}
where $\bm f_n = f_n \, \bm n_i$ and  $\bm f$ is the force measured at the tool reference frame (using the joint torque sensors). The final
actuator torque command $\bm \tau$ that is sent to the robot is the summation of (\ref{eqn:tau_imp}) and (\ref{eqn:tau_n}).

\begin{comment}
An integral image IO corresponding to an image O makes
it possible to compute the sum of all values of O within
a certain rectangular region R(ms,ns)->(me,ne)

To perform this averaging integral images
is performed.
It consist in create two maps of tangential vectors, one for the u- and one for the v-direction. 
The vectors for these maps are computed between corresponding 3D points in the point cloud. 
That is, each element of theses vectors is a 3D vector. For each of the channels (Cartesian x, y, and z)
of each of the maps we then compute an integral image, which leads to a total
number of six integral images. Using these integral images, we can compute the
average tangential vectors with only 2 × 4 × 3 memory accesses, independent of
the size of the smoothing area. The overall runtime complexity is linear in the
number of points for which normals are computed     

... 
\end{comment}

\section{Case study: Demonstration at Kuka Innovation Award Competition 2018}
        \label{sec:experiments}

\textbf{Demo Setup} As the final stage of the KUKA innovation award competition, we presented our system at Hanover Messe. It was in the format of a live demo, running for five days, 6 hours per day. 
The algorithm of the demo system ran on the following hardware.
The user interface ran on a Thinkpad X230i touch-screen laptop. 
The backend system ran on two commodity computers, both of which had a Quad-core CPU running at 4GHz.
One of them was for the message passing of the sensors and the processing of the point cloud. The other ran the incremental learning algorithms on a NVIDIA GTX 960.
Despite the moderate computing power of the GPU, we could get near real-time performance from our localization and recognition algorithm, thanks to the optimization described in Section~\ref{sec:localization}. 
The inference time of our incremental detection model was on average 200ms per image, accounting for both the localization and the recognition. 

\begin{figure}[h]
\centering
\includegraphics[trim=0mm 30mm 0mm 05mm, clip=true, width=.3\textwidth] {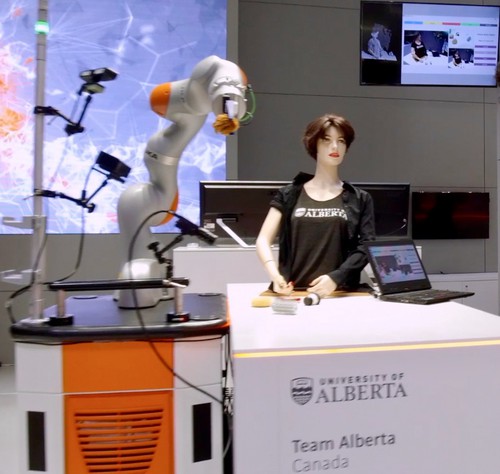}
\caption{Team Alberta booth at Hannover Messe}
\label{fig:team}
\end{figure}
        
The vision sensors included a pointgrey camera for  teaching  objects,  an asus-X  RGB-D  sensor for extracting the 3D point cloud of the surface, a Kinect2 sensor for object detection and mapping 2D localization result to 3D. 
Both of the RGB-D sensors were calibrated with respect to the robot base frame. 
The accuracy of the surface reconstruction is limited by the resolution of the RGBD-sensor, in our case $\pm 5^{mm}$ within the workspace of the robot, measured by moving the end-effector toward a target that the user specifies through the user interface.
In other words, the error generated by the 2D-3D projection of the user-defined path is $\pm 5^{mm}$.
The compliance controller would be able to deal with the inaccuracy and ensure the continuous contact during the motion.

When teaching the object, we cropped the image from pointgrey camera to 400 by 400 pixels and resized to 224 by 224 pixels before feeding into the network.
The input to object detection module was the HD image stream from the Kinect2 sensor. Again, it was cropped to 600 by 600 pixels, before resized to 224 by 224 pixels, for a good balance of speed and performance.

What was unique about our live demo in contrast to a traditional lab setup, was that there was no direct control over the interaction with the audience nor the lighting conditions, which changed due to overhead sky-lights. 

A full run of each competition demo was required by the organizers to run in a tight four-minute window, consisting of two phases, (1) a new object detection phase, and (2) a robot motion teaching phase with the new object. In phase one, the image collection of new object took
40 seconds plus another 40 seconds for deep features extraction. Presenting the object detection results to the audience took another 30 seconds.  In phase two, visual motion teaching on the laptop touch screen took 10 seconds, and path planning 20 seconds. Finally autonomous execution of the surface contact motion took one and a half minute (the robot manipulator performed a reaching and grasping motion, then followed the trajectory defined on the unstructured surface). These were the average duration of each phase. The specific duration varied from demo to demo.

In order to test the limits of our system, 
we encouraged the visitors to bring their own objects and teach the robot.
In Table~\ref{tab:1}, we summarized the classification result of the new objects brought by the visitors at the demo.

\begin{table}[h]
    \centering
        \caption{List of new objects tested by the audience. The unsuccessful classification refers to the cases where the new object was localized but recognized as an ``Unknown'' category.}
    \begin{tabular}{l|l}
        Successful classification $ \checkmark $         & Unsuccessful classification $ \times $ \\
        \hline
        cellphone       &   lanyards            \\
        wallets         &   earPods             \\
        sponge ball     &   cables              \\
        plush toys            &   brochure            \\
        key chains      &    car key         \\
        water bottle    &               \\
        fidget spinner  &               \\
        pen & \\
        business card & \\
        card holder &  \\
        lighter & 
        \end{tabular}
    \label{tab:1}
\end{table}
\textbf{Discussion}
Overall, the system performed robustly even in the presence of many kinds of uncertainties. 
For example, a big challenge was to deal with lighting variations. There was an opening on the ceiling near the setup, causing a mixture of natural and indoor lighting that changed dramatically during the day. 
As shown in Table~\ref{tab:1}, the incremental learning algorithm classified the object successfully on about 69\% of all the object categories introduced by the audience.
Of the failure cases, most were due to the following characteristics of the objects:
flexible material, tiny shape, reflective surface, similar textures and colors to the objects in the training set.
In total, we performed about 75 demos where about 65 were successful in the incremental learning part, which came to 87\% success rate of the demos.
Note that the running time of the incremental classification algorithm grows linearly with the total number of existing classes, due to the nearest neighbor approach. To ensure the speed of the demos, we reset the system to five classes at the end of each day.

In the motion execution phase, the contact motions were always successful as long as the defined path was within reach of the manipulator. 
Since the robot applies force to maintain the contact with the mannequin, there may be a concern whether it may be safe or comfortable for a real human. In fact, it is already addressed in our algorithm. The desired force at the end effector is specified in the program, which transfers to a desired torque that gets sent to the robot. This force can be set to a value that falls in the safe range described in ISO 15066.
In the future, we can easily modify the visual interface to allow the operator to adjust the desired contact force based on the target application.

The unsuccessful cases were either the gripper failed to pick up the object, or the defined path was not reachable.
We provided failure recovery mode for each case.
In the case of a pick up failure, the user could stop the motion of the robot by pressing the button on the robot flange and manually pass the object to the gripper.
In the case of non-reachable path, the robot would signal the user such that he/she can redefine the path.
As the focus of our work was not on grasping, a simple table-top grasping was considered for the objects based on the shape of the bounding box. Obviously we had limitations due to the nature of the provided  gripper and we were not able to grasp objects that were too large (greater than $10^{cm}$ in diameter) or too small (less than $1^{cm}$).

\section{Conclusions}
        \label{sec:conclusions}
        
This paper introduced a novel system that  acquired knowledge incrementally from human, where new tools and motions can be taught on the fly. 
The usage of the system was showcased in live demonstrations as the final round  of the KUKA Innovation Award competition.  
Our system made it possible for users to define new classes of object that could be recognized later. It also  allowed the user  to associate specific tasks  for these new objects and perform actions with them. 

At the demos, our system has attracted the attention from multiple companies for possible trials, including bin-picking applications where they were specifically interested in fast teaching of new objects. 
Also, there were interests from cosmetic industry for 3D path specification and automotive industry 3D contour inspection.

\addtolength{\textheight}{-12.5cm}

\bibliographystyle{IEEEtran}
\bibliography{Ref}

\end{document}